\documentclass[conference]{IEEEtran}
\IEEEoverridecommandlockouts
\usepackage{cite}
\usepackage{amsmath,amssymb,amsfonts}
\usepackage{algorithmic}
\usepackage{graphicx}
\usepackage{textcomp}
\usepackage{algorithm}
\usepackage{float}
\usepackage{tikz}
\usepackage{pgf-pie}
\usepackage{multirow}
\usepackage{hhline}
\usepackage{subcaption}

\def\BibTeX{{\rm B\kern-.05em{\sc i\kern-.025em b}\kern-.08em
    T\kern-.1667em\lower.7ex\hbox{E}\kern-.125emX}}
\begin{document}

\title{Efficient Single-Shot Multibox Detector for Construction Site Monitoring}


\author{\IEEEauthorblockN{1\textsuperscript{st} Viral Thakar}
\IEEEauthorblockA{\textit{Electrical and Computer Engineering} \\
\textit{Concordia University}\\
Montreal, Canada \\
v\_thakar@encs.concordia.ca}
\\
\IEEEauthorblockN{4\textsuperscript{th} Mohammad M Soltani}
\IEEEauthorblockA{
\textit{Indus.ai}\\
Thornhill, Canada \\
mohammad.soltani@indus.ai}
\and
\IEEEauthorblockN{2\textsuperscript{nd} Himani Saini}
\IEEEauthorblockA{\textit{Computer Science and Software Engineering} \\
\textit{Concordia University}\\
Montreal, Canada \\
h\_ain@encs.concordia.ca}
\\
\IEEEauthorblockN{5\textsuperscript{th} Ahmed Aly}
\IEEEauthorblockA{
\textit{Indus.ai}\\
Thornhill, Canada \\
ahmed.aly@indus.ai}
\and
\IEEEauthorblockN{3\textsuperscript{rd} Walid Ahmed}
\IEEEauthorblockA{\textit{CIISE} \\
\textit{Concordia University}\\
Montreal, Canada \\
walidmohamed.ahmed@concordia.ca}
\\
\IEEEauthorblockN{6\textsuperscript{th} Jia Yuan Yu}
\IEEEauthorblockA{\textit{CIISE} \\
\textit{Concordia University}\\
Montreal, Canada \\
jiayuan.yu@concordia.ca}
}

\maketitle
\begin{abstract}
Asset monitoring in construction sites is an intricate, manually intensive task, that can highly benefit from automated solutions engineered using deep neural networks. We use Single-Shot Multibox Detector --- SSD, for its fine balance between speed and accuracy, to leverage ubiquitously available images and videos from the surveillance cameras on the construction sites and automate the monitoring tasks, hence enabling project managers to better track the performance and optimize the utilization of each resource. We propose to improve the performance of SSD by clustering the predicted boxes instead of a greedy approach like non-maximum suppression. We do so using Affinity Propagation Clustering --- APC to cluster the predicted boxes based on the similarity index computed using the spatial features as well as location of predicted boxes. In our attempts, we have been able to improve the mean average precision of SSD by 3.77\% on custom dataset consist of images from construction sites and by 1.67\% on PASCAL VOC Challenge. 
\end{abstract}

\section{Introduction}
Construction sites forms an important parts of cities: they are intricate environments with a broad range of activities like clearing, dredging, excavating, and building \cite{batty2012smart}. 
These activities require large numbers of expensive equipments, and it is crucial to monitor their proper utilization. This monitoring is time-consuming, labor intensive, and prone to human errors by project managers. Smart Cities incorporate a large number of Internet-connected sensors and actuators. This paper considers the use of IP cameras such as sensors to automate monitoring on construction sites.  

We propose an automated system to detect, localize and classify equipment from videos to generate real-time reports that facilitate decision-making. We do so using recently introduced computer vision algorithms \cite{lecun2015deep,voulodimos2018deep} trained on surveillance videos from construction sites that are available but underutilized in computer vision research. We use Single-Shot MultiBox Detector --- SSD with Non-Maximum Suppression --- NMS \cite{DBLP:journals/corr/LiuAESR15} as base model, which reaches record performance for object detection, scoring over 74\% mAP (mean Average Precision) at a real-time rate of 59 frames per second on the Pascal VOC Challenge \cite{everingham2010pascal}. 

\begin{figure*}
	\centering
    \begin{subfigure}[b]{0.30\linewidth}
        \includegraphics[width=\textwidth]{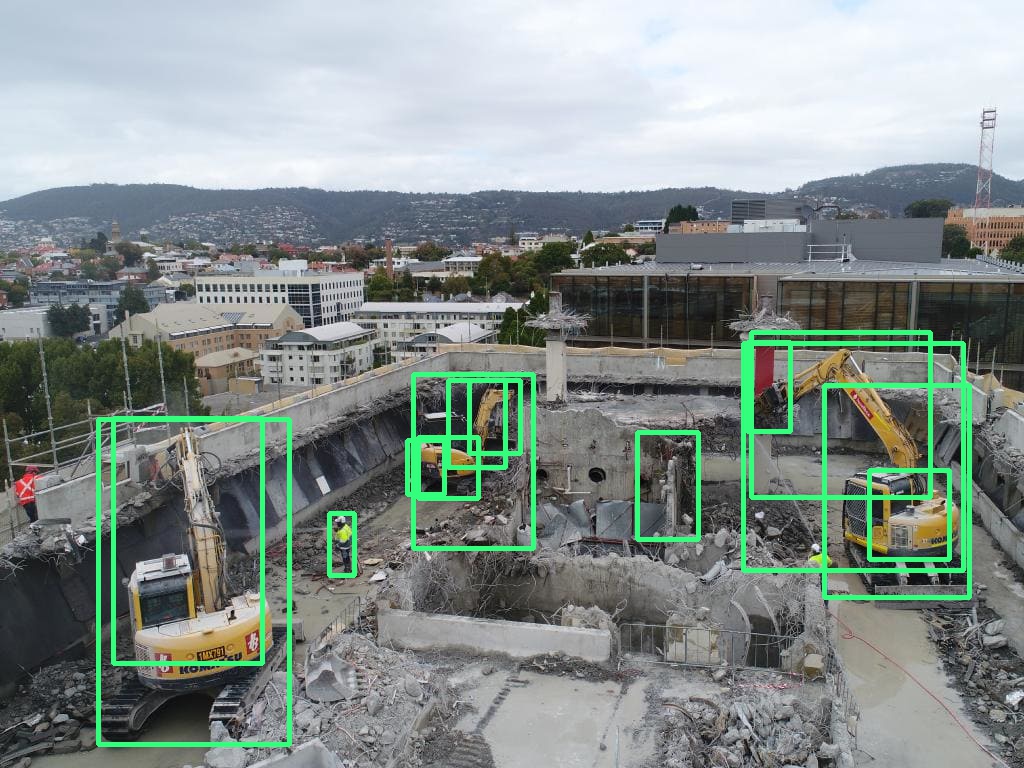}
        \caption{Predicted Boxes by SSD without NMS}
        \label{fig:ssdwonms}
    \end{subfigure}
    ~
    \begin{subfigure}[b]{0.30\linewidth}
        \includegraphics[width=\textwidth]{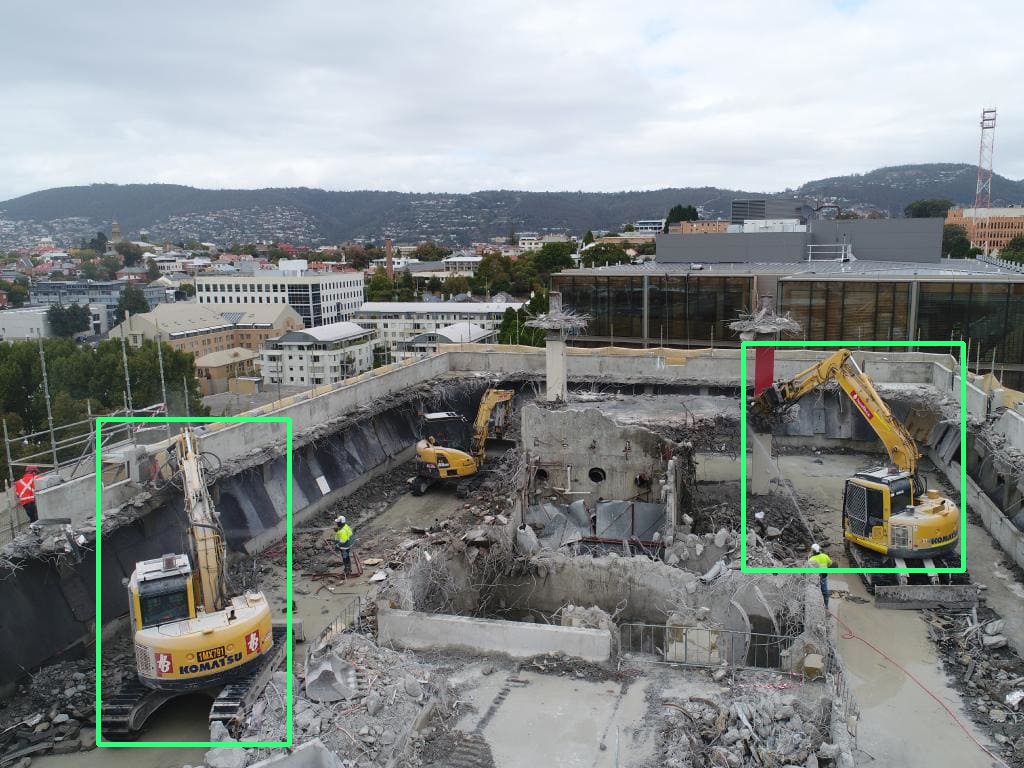}
        \caption{Predicted Boxes by SSD with NMS}
        \label{fig:ssdnms}
    \end{subfigure}
    ~
    \begin{subfigure}[b]{0.30\linewidth}
        \includegraphics[width=\textwidth]{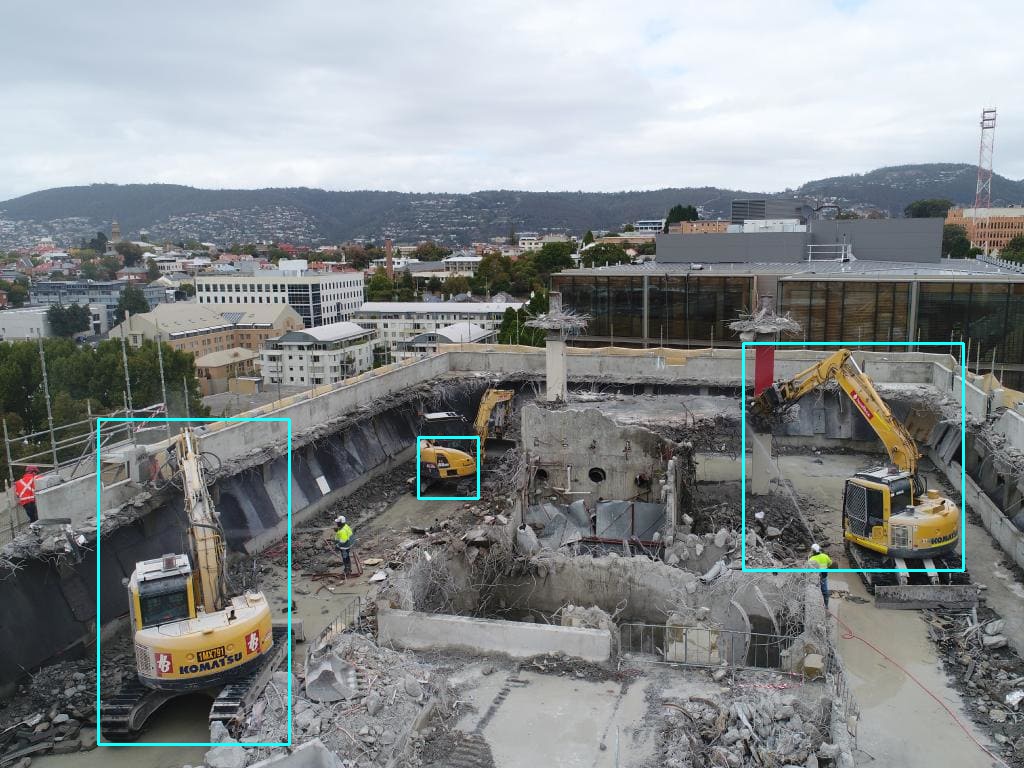}
        \caption{Predicted Boxes by SSD with APC}
        \label{fig:ssdapc}
    \end{subfigure}
    \caption{Comparison of SSD with NMS and with APC}\label{fig:comparison}
\end{figure*}

SSD uses greedy NMS, where out of all the detected bounding boxes, the boxes with higher confidences are selected and the other boxes overlapping the selected boxes are suppressed subjected to an intersection over union ($iou$) threshold. NMS uses a static threshold, usually $0.5$, to winnow away candidate bounding boxes. But this very technique of NMS causes the detector to fail while looking for objects which appear smaller or have low resolution because of far away camera placement on construction sites. Fig.\ref{fig:ssdwonms} and Fig.\ref{fig:ssdnms} show this particular problem where SSD is not able to detect a small equipment.  

We propose to replace NMS with Affinity Propagation Clustering \cite{frey2007clustering} to solve the drawback of greedy NMS. We choose APC over other clustering techniques, because it does not require the number of clusters to be determined a priori. Rather, it is an exemplar based algorithm that employs a simple message-passing technique to cluster based on similarity between bounding boxes generated by SSD. The algorithm progressively engenders communication between each bounding box and its exemplar to produce a high-quality set of clusters with the exemplars as the cluster centers. 

In our technique, we cluster the bounding boxes on the basis of the location as well as the appearance-based spatial features of the pixels enclosed by those bounding boxes, hence distinguishing the objects based on both their location as well as appearance. This allows the selection of final detection to be adaptive in contrast to the hand-designed threshold values of NMS, and since there is no minimum value for the detection overlap, small objects are also detected successfully. Fig.\ref{fig:comparison} depicts the performance comparison of SSD with NMS and with APC. 

The paper is organized as follows. Section 2 describes the SSD and APC to provide intuition of the problem and capability of clustering to solve it. It also introduces the notation used in the paper. Section 3 describes the proposed algorithm. Section 4 covers the dataset used, experimental setup and result analysis under different situations. 

\section{Preliminaries}
\subsection{Single Shot Multibox Detector}

The key feature of Single Shot Multibox Detector --- SSD \cite{DBLP:journals/corr/LiuAESR15} is a feed-forward convolutional network, that can, in a single pass perform both the object classification, by predicting the class score and object localization, by performing bounding box regression followed by non-maximum suppression --- NMS. Let $\chi$ be the domain set representing all the images from which  objects are to be detected and $M = \{1, 2, ... l\}$ be the label set representing the labels for all the object-classes in those images. Let $D = \{d_1, d_2, ...d_n\}$ be the set of default boxes created using Algorithm 1 over different aspect ratios and regularly spaced scales. Each default box $d_i$ is a vector representing four values associated with the box $[c_x\ c_y\ w\ h]$, where $(c_x, c_y)$ are the coordinates of the centroid and $w$ and $h$ are respectively the width and height of the bounding box. SSD is a function $\varphi(x) = \hat{Y}$ that takes an arbitrary image $x \in \chi$ as an input and produces a matrix $\hat{Y} \in \mathbb{R}^{n \times (l + 4)}$ as the output. Each row of $\hat{Y}$ represents a $(l + 4)$ dimensional positive real valued vector, which contains $l$ per-class classification probabilities or confidences and four offsets in the default box dimensions. For simplicity consider $\hat{Y} = [\hat{Z}\ \hat{B}]$ where matrix $\hat{Z} \in \mathbb{R}^{n \times l}$ represents classification task and $\hat{B} \in \mathbb{R}^{n \times 4}$ represents localization task. 

The training data available for object detection consists of images and the boxes circumscribing the objects (called ground-truth boxes) along with their class labels. Hence, for any arbitrary image $x \in \chi$, $G = \{g_1, g_2, ... g_h\}$ represents a set of ground-truth boxes, where $g_i \in \mathbb{R}^{(l+4)}$. The training objective of SSD is to learn the prediction rules to predict the object class present in each default box and the amount of offset required in the shape of the default boxes with respect to the ground-truth box. This correlation, associated with each default box is computed using Algorithm 2 and stored in a ground-truth matrix $Y \in \mathbb{R}^{n \times (l + 4)}$.

\begin{algorithm}
   \caption{Initialize Set of Default Boxes $D$}
   \label{alg1}
\begin{algorithmic}
	\STATE{\bfseries Inputs:}
   		\\$p$ - Number of feature maps
        \\$f \in \mathbb{R}^p$ where $\forall k \in [1,p] : f[k]$ - Dimension of $k^{th}$ square feature map
        \\$s_{min}$ - Minimum Scale Value - Default 0.2
        \\$s_{max}$ - Maximum Scale Value - Default 0.9
    \STATE{\bfseries Initialize:}
    	\\$D = \{\}$ - Set of Default Boxes
    \STATE{\bfseries Process:}
    	\FOR{$k \in [1, p]$}
          \STATE{Set of centroids of default boxes in $k^{th}$ feature map}   
              	\\$$\ \forall i,j \in [0, f[k]): C_k = \left\{\left(\frac{i + 0.5}{f[k]}, \frac{j + 0.5}{f[k]}\right)\right\}$$
          \STATE{Scale values of default boxes in $k^{th}$ feature map}
          		\\$$s_k = s_{min} + \frac{s_{max} - s_{min}}{p - 1}(k - 1),\ s'_k = \sqrt[]{s_k \cdot s_{k+1}}$$
          \STATE{Compute Default boxes for $k^{th}$ feature map}   
          	\FOR{$\forall(c_x, c_y) \in C_k, \forall a \in \{2, 3, 1/2, 1/3 \} $:}
              \STATE{\begin{align*}
              		D &= D \bigcup\ \{[c_x,\ c_y,\ s_k \sqrt[]{a},\ s_k / \sqrt[]{a}]\} 
              \end{align*}}  
            \ENDFOR
              
		 \STATE{$D = D \bigcup \{[c_x,\ c_y,\ s_k,\ s_k], [c_x,\ c_y,\ s'_k,\ s'_k]\}$}
       \ENDFOR
	\STATE{\bfseries Output:}
		\\$D$ - Set of Default Boxes
            
\end{algorithmic}
\end{algorithm}

\begin{algorithm}
   \caption{Create Ground Truth Matrix $Y$}
   \label{alg2}
\begin{algorithmic}
	\STATE{\bfseries Inputs:}
   		\\$\tau$ - Overlap threshold - Default 0.5
        \\$D = \{d_1, ..., d_n\}$ - Set of Default Boxes
        \\$G = \{g_1, ..., g_h\}$ - Set of Ground-truth Boxes
	\STATE{\bfseries Initialize:}
    	\\$Y \in \mathbb{R}^{n \times (l + 4)}\ such\ that\ Y = [Z\ B] = [0]$
        \\$pos, neg = \{\}$ - Sets to store indexes of positively and negatively matched default boxes
    \STATE{\bfseries Procedure:}
    	\FOR{$i \in [1, n]$}
        	\FOR{$j \in [1, h]$}
            	\STATE{$g = g_j$ and $d = d_i$}
                \STATE{$iou = 1 - \frac{|g_{l+1 : l+4} \cap d|}{|g_{l+1 : l+4} \cup d|}$}
            	\IF{$iou \geq \tau$}
                	\STATE{$\mathrm{class} = \arg\max(g_{1:l})$}
                    \STATE{$\forall c \in class : Z_{i, c} = 1$}
                    \STATE{$B_{i,:} = g_{l+1:l+4}$}
                    \STATE{$pos = pos \cup \{i\}$}
                \ELSE
                	\STATE{$neg = neg \cup \{i\}$}
                \ENDIF
            \ENDFOR
        \ENDFOR
	\STATE{\bfseries Output:}
    	\\$Y = [Z\ B]$ 
\end{algorithmic}
\end{algorithm}

SSD uses weighted sum of classification and localization loss as an overall loss function, minimize it using Adam Optimizer \cite{kingma2014adam} during the training. For classification, SSD calculates the multi-class softmax loss. If $N$ is the total number of positively matched default boxes in Algorithm 2 and $\sigma \in M$ is some label representing the background, the classification loss is:
\begin{equation}
L_{clss}(\hat{Z}, Z) = - \sum_{c=1}^{l} \sum_{i \in pos} Z_{i,c} \cdot \log(\hat{Z}_{i,c}) - \sum_{j \in neg} \log(\hat{Z}_{j,\sigma})
\end{equation}
Let
$$smooth_{L_1}(x) = \Big\{\begin{tabular}{c} $0.5 \times x^2$ if $|x| < 1$ \\ $|x| - 0.5$ otherwise \end{tabular}$$
denote the smooth L1 function \cite{DBLP:journals/corr/Girshick15}.
The localization loss is:
\begin{equation}
L_{loc}(\hat{B}, B) = \sum_{i \in pos} smooth_{L_1}(\hat{B}_{i,:} - B_{i,:})
\end{equation}
Hence, the overall loss is: 
\begin{equation} 
L(\hat{Y}, Y) = \frac{1}{N} \Big(L_{clss}(\hat{Z}, Z) + \alpha L_{loc}(\hat{B}, B)\Big)
\end{equation}
where $\alpha$ is the weight value that controls the balance between the two losses.  

The localization and classification tasks are followed by two post-processing steps. First, SSD creates a matrix $\Omega \in \mathbb{R}^{n \times (l+4)}$ representing an ordered set of predicted boxes given as: 
\begin{equation}
\Omega = [\hat{Z}\ \hat{B}+D]
\end{equation}
Each row of matrix $\Omega$ represents the $l$ class probabilities with the exact bounding box circumscribing the object. Secondly, SSD performs per class non-maximum suppression to produce the final detection. For that, it removes all the predicted boxes which belong to the background class; and then, it iteratively performs non-maximum suppression for the other classes by selecting the most confident predicted box, and removing all other overlapping boxes with $iou > 0.5$, till there is no box overlapping the selected one. 
 
\subsection{Affinity Propagation Clustering}
Affinity Propagation Clustering clusters the data by exchanging certain real-valued similarity messages between the pairs of data points until convergence, producing a refined set of clusters and corresponding exemplars.

APC takes as inputs a set of data points $T = \{t_1, t_2, ... t_q\}$ and a similarity matrix  $S \in \mathbb{R}^{q \times q}$, whose element  $S_{i, j}$ is a measure of the similarity between data points $t_i$ and $t_j $, computed as per function $s(t_i, t_j)$. Additionally, it takes a preference vector $\rho  = \{\rho_1, \rho_2, ... \rho_q\}$ , where $\rho_i$ is associated with each data point $t_i$ such that $t_i$ with larger $\rho_i$ are more likely to be the  exemplars. The output of APC is largely influenced by the choice of  $\rho$, as in, choosing a shared value (e.g. median) can result in moderate number of clusters, whereas choosing a minimum value can result in a small number of clusters.
 
Two types of messages are iteratively exchanged between the node pairs, that can be combined at any iteration to give the clusters and their respectively chosen exemplars. A \textit{responsibility} message $r(i, j) \in \mathbb{R}$ from the data point  $t_i$ to a potential exemplar $t_j$, reflecting the suitability of  $t_j$ to be an exemplar for  $t_i$, given the other potential exemplars; and  an \textit{availability} matrix  $a(i, j) \in \mathbb{R}$ from a candidate exemplar  $t_j$ to the data point  $t_i$ reflecting how appropriate it would be for  $t_j$ to serve as an exemplar of  $t_i$, given the support from the other data points for  $t_j$ to be an exemplar. In each iteration, these messages are updated as per Algorithm 3, until either the changes in the messages fall below some threshold, or there is no update in the computed clusters and  corresponding exemplars over some iterations.

\begin{algorithm}
   \caption{Affinity Propagation Clustering}
   \label{alg4}
\begin{algorithmic}[1]
	\STATE{\bfseries Inputs:}
   		\\$T = \{t_1, t_2, \ldots, t_q\}$ - Set of data points
        \\$\rho \in \mathbb{R}^q$  - Preference vector
        \\$S$ - Set of pairwise similarities 
        $$\forall (i, j) \in \{1, \ldots, q\}^2  : S_{i,j} = \begin{cases}s(t_i, t_j),\ i \neq j\\\\
        					\rho_j,\ i = j
        		\end{cases}$$    
   	\STATE{\bfseries Initialization:}
    	\\$\forall (i, j) \in \{1, 2, ... q\}^2 : a(i, j) = 0$
    \STATE{\bfseries Repeat until convergence:}
    \begin{align*}
    \forall(i, j) &\in \{1, 2, ... q\}^2 :\\
    r(i, j) &= S_{i,j} - \max_{k : k \neq j}[S_{i,k} + a(i, k)]\\
    a(i, j) &= \begin{cases}
               							\sum_{k : k \neq i} \max[0, r(k, j)],\quad \mbox{if }j = i\\\\
               							\min[0, r(j, j) + \sum_{k : k \notin \{i, j\}} \max[0, r(k, j)]], \mathrm{ow}
            							\end{cases}
    \end{align*}

    \STATE{\bfseries Output:}
    	\\Assignment vector $\hat{c} \in \mathbb{R}^q$ such that 
        \\$$\forall i \in [1,q], \forall j \in [1,q] : \hat{c}_i = \operatorname*{argmax}_j [a(i, j) + r(i, j)]$$
        \\Set of exemplar data points
        \\$$\forall i \in [1, q] : E = {T[\hat{c}_i]}$$
    
\end{algorithmic}
\end{algorithm}

\section{Proposed Architecture}
The proposed object detection pipeline for SSD with APC is shown in figure. To produce final predictions, instead of applying non-maximum suppression to the boxes predicted by SSD, we propose to cluster them based on their similarity using APC. We compute the preference vector $\rho$ of APC from the predicted matrix $\Omega \in \mathbb{R}^{n \times (l+4)}$ of SSD as $\forall i \in [1,n] : \rho_i = \max[\Omega_{i,1:l}]$. This allows APC to select boxes with high predicted confidences as exemplars. The similarity between two predicted boxes is calculated as a weighted sum of the location of the default box and its computed visual based features. The location based similarity is computed as $iou$ between two predicted boxes. The visual appearance based similarity is computed as euclidean distance between histogram of gradients \cite{dalal2005histograms} computed for those two predicted segments of the image. 

Consider the label set $M = \{1, 2, ... l\}$, where 1 indicates the background and  $\{2, 3 ... l\}$ indicates the labels for different object classes. For an arbitrary image $x$ with ground-truth boxes $G = \{g_1, g_2, ... g_h\}$, the predicted output of SSD $\Omega \in \mathbb{R}^{n \times (l + 4)}$ represents the $n$ segments of input image. Let $\Delta \in \mathbb{R}^{n \times 4}$ where $\forall i \in [1, n] : \Delta_{i,:} = \Omega_{i, l+1:l+4}$ represents the predicted bounding boxes. The location-based similarity can be calculated as: 
\begin{equation}
\forall (i, j) \in \{1, 2, ... n\}^2, i \neq j  : \alpha_{i,j} = 1 - \frac{|\Delta_{i,:} \cap \Delta_{j,:}|}{|\Delta_{i,:} \cup \Delta_{j,:}}.\end{equation}
Let $\forall i \in [1,n] : \eta_i$ represents the histogram of gradient feature vector calculated using the method proposed in \cite{dalal2005histograms}. The visual appearance based similarity can be calculated as following:
\begin{equation}
\forall (i, j) \in \{1, 2, ... n\}^2, i \neq j  : \beta_{i, j} = - ||\eta_i - \eta_j||^2.
\end{equation}

For real-valued weight factor $\lambda \in [0,1]$, the elements of the similarity matrix $S$ can be computed as: 
\begin{equation}
\forall (i, j) \in \{1, 2, ... q\}^2  : S_{i,j} = \begin{cases}\frac{\alpha_{i,j} + \lambda \beta_{i,j}}{2},\ i \neq j\\
        					\rho_i,\ i = j
        		\end{cases}.
\end{equation}
Using similarity matrix from equation 7 with Algorithm 3 gives $E$, the set of exemplars of the  predicted boxes representing the final detections of the objects. Ideally this $E$ should be close to the ground-truth $G$.



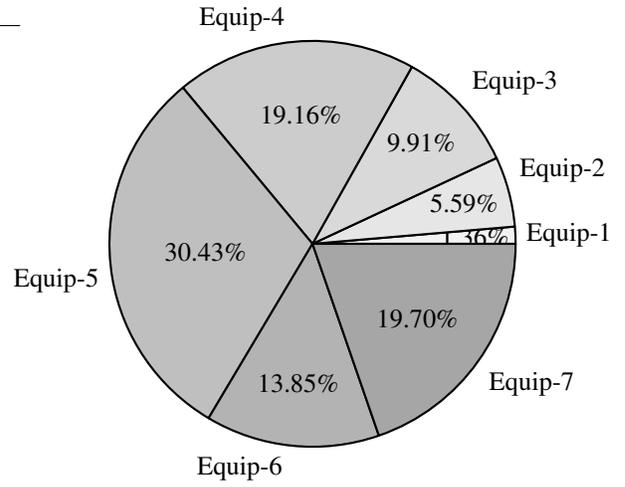
\begin{figure}
\begin{tikzpicture}
\pie [radius=2.7, color={black!5, black!10, black!15, black!20, black!25, black!30, black!35}]{1.36/Equip-1, 5.59/Equip-2, 9.91/Equip-3, 19.16/Equip-4, 30.43/Equip-5, 13.85/Equip-6, 19.70/Equip-7} 
\end{tikzpicture}
\caption{Training Dataset Statistics}\label{fig:traindatastats}
\end{figure}

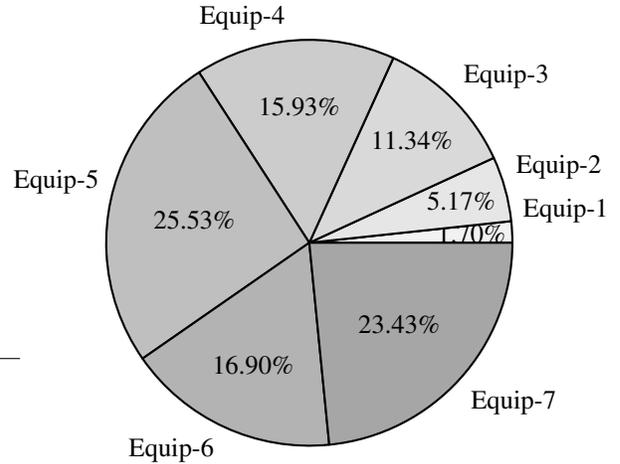
\begin{figure}
\begin{tikzpicture}
\pie [radius=2.7, color={black!5, black!10, black!15, black!20, black!25, black!30, black!35}]{1.70/Equip-1, 5.17/Equip-2, 11.34/Equip-3, 15.93/Equip-4, 25.53/Equip-5, 16.90/Equip-6, 23.43/Equip-7} 
\end{tikzpicture}
\caption{Testing Dataset Statistics}\label{fig:testdatastats}
\end{figure}

\begin{figure}
	\includegraphics[width=\linewidth, height=4cm]{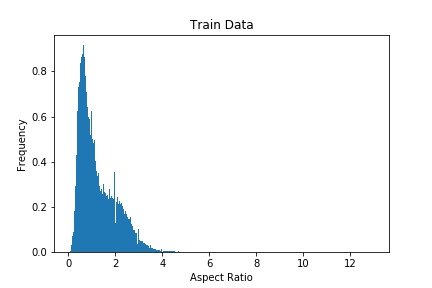}
    \includegraphics[width=\linewidth, height=4cm]{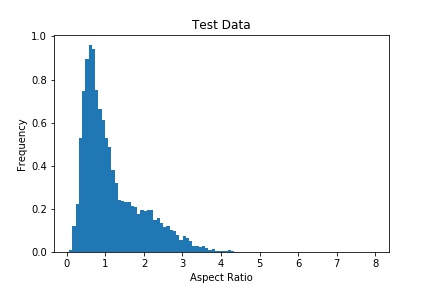}
    \caption{Aspect Ratio Distribution of Train and Test Data}\label{fig:ardistr}
\end{figure}

\begin{table*}
\centering
\caption{Average Precision (\%) per Class and Mean Average Precision (\%) Comparison for Custom Dataset}
\label{customdata}
\begin{tabular}{|c|c|c|c|c|c|c|c|c|c|c|}
\hline
\multirow{1}{*}{\textbf{Method}} &
\multirow{1}{*}{\textbf{Time (ms)}} &
\multirow{1}{*}{\textbf{Equip-1}} & \multirow{1}{*}{\textbf{Equip-2}} & \multirow{1}{*}{\textbf{Equip-3}} & \multirow{1}{*}{\textbf{Equip-4}} & \multirow{1}{*}{\textbf{Equip-5}} & \multirow{1}{*}{\textbf{Equip-6}} & \multirow{1}{*}{\textbf{Equip-7}} & \multirow{1}{*}{\textbf{mAP (\%)}} & \multirow{1}{*}{\textbf{\% Improvement}}\\ \hline \hline

\textbf{SSD-MobileNet} & 32 & 59.12 & 70.45 & 44.62 & 69.41 & 15.50 & 53.14 & 48.56 & 51.24 & \multirow{2}{*}{\textbf{3.67 \%}}\\ 
\textbf{SSD-MobileNet-APC} & 32 & 63.40 & 72.89 & 47.32 & 72.40 & 18.25 & 58.14 & 54.13 & 55.21 & \\ \hline
\textbf{SSD-Inception} & 45 & 75.56 & 80.74 & 57.45 & 79.19 & 24.58 & 70.09 & 58.39 & 63.76 & \multirow{2}{*}{\textbf{3.86 \%}}\\
\textbf{SSD-Inception-APC} & 46 & 79.12 & 83.80 & 60.01 & 82.11 & 29.34 & 74.27 & 64.23 & 67.55 & \\ \hline
\end{tabular}
\end{table*}

\begin{table}
\centering
\caption{Average Precision per Class and Mean Average Precision Comparison for PASCAL VOC}
\label{pascaldata}
\begin{tabular}{|c|c|c|}
\hline
\multirow{1}{*}{\textbf{Class}} & \multirow{1}{*}{\textbf{SSD-VGG16}} & \multirow{1}{*}{\textbf{SSD-VGG16-APC}} \\ \hline
Aeroplane	& 75.5	& 76.1 \\ \hline
Bicycle		& 80.2	& 82.3 \\ \hline
Bird		& 72.3	& 73.5 \\ \hline
Boat		& 66.3	& 68.2 \\ \hline
Bottle	& 46.6	& 48.7	  \\ \hline
Bus	& 83.0	& 85.12 \\ \hline
Car			& 84.2	& 84.3 \\ \hline
Cat & 86.1	& 88.3	\\ \hline
Chair & 54.7	& 56.6	\\ \hline
Cow & 78.3	& 79.3	\\ \hline
Dinning Table & 73.9	& 76.2	\\ \hline
Dog & 84.5	& 85.2	\\ \hline
Horse & 85.3 & 85.5 \\ \hline
Motor bike & 82.6	& 83.3	\\ \hline
Person & 76.2	& 79.0	\\ \hline
PottedPlant & 48.6	& 49.6	\\ \hline
Sheep & 73.9	& 76.5	\\ \hline
Sofa & 76.0	& 77.2	\\ \hline
Train & 83.4	& 85.9	\\ \hline
TV & 74.0	& 76.2	\\ \hline
mAP	(\%)		& 74.3	& 75.9 \\ \hline
Time (ms)	& 38 & 38 \\ \hline
\% \textbf{Improvement} & \multicolumn{2}{|c|}{\textbf{1.6}} \\ \hline
\end{tabular}
\end{table}

\section{Experimental Results and Comparison}
As the goal of this research is to provide a better object detection algorithm to develop an automated asset monitoring and management system for the construction sites, we have created the dataset from images and videos, captured at different construction sites, having a wide range of equipments. The dataset has seven labels : Equipment-1 to Equipment-7. The images are taken from surveillance IP cameras placed at various construction sites with different angles and heights. This setup allows us to generate a dataset with objects ranging in different scales as well as aspect ratios. Figure \ref{fig:traindatastats} and \ref{fig:testdatastats} shows the details about training and testing dataset. Figure \ref{fig:ardistr} shows the distribution of aspect ratios of objects present in the dataset. We also test the proposed approach on PASCAL VOC dataset \cite{everingham2010pascal} for fair comparison with SSD architecture proposed in \cite{DBLP:journals/corr/LiuAESR15}. The training and testing distribution of PASCAL VOC dataset is also the same as SSD-300 architecture in \cite{DBLP:journals/corr/LiuAESR15}. The performance evaluation of both the datasets is done and analyzed separately.      

To understand the performance, we consider three variants of SSD based on different feature extraction network. i) SSD with Inception \cite{DBLP:journals/corr/SzegedyVISW15} ii) SSD with Mobilenet \cite{DBLP:journals/corr/HowardZCKWWAA17} and iii) SSD with VGG-16 \cite{DBLP:journals/corr/LiuAESR15,simonyan2014very}. These different feature extraction networks cover almost all the state-of-the-art variants of SSD available and allow us to verify effectiveness of the proposed algorithm on different architectures. We evaluate the PASCAL VOC evaluation metrics and COCO evaluation metrics to compare performance of proposed algorithm. 

\subsubsection{Evaluation on Custom Dataset}
At first, we train SSD-Inception and SSD-Mobilenet with the custom dataset from construction sites. We have two versions of each architecture, one with NMS (SSD-Mobilenet and SSD-Inception) and another created using proposed algorithm (SSD-Mobilenet-APC and SSD-Inception-APC). We evaluate all the models for per-class average precision as well as mean average precision as suggested in PASCAL VOC evaluation metrics \cite{everingham2010pascal}. The evaluation of these models provide us an insight about the performance of all the variants of SSD on different custom object classes. The results in Table 1 indicate the per class average precision comparison of conventional SSD with NMS against proposed SSD with APC for each variant. 

We observed that conventional SSD is fairly able to detect large objects, as in objects covering larger area with respect to total area of the image but, struggles against smaller objects. On the other hand, performance of SSD with APC is far better for smaller objects. It achieves this by detecting objects, which are rejected by conventional SSD during the NMS process. 



To verify the authenticity of proposed improvement and provide fair evaluation, we compared the conventional SSD model provided in \cite{DBLP:journals/corr/LiuAESR15} with the proposed algorithm on PASCAL VOC dataset. For this evaluation we used SSD with VGG-16 as base network. We kept all the default values as well as hyper parameters of the network same as the values mentioned in \cite{DBLP:journals/corr/LiuAESR15}. Table 2 provides the comparison for per class average precision and mean average precision. The real-time detection is the main reason of selecting SSD as base algorithm and replacing NMS with APC doesn't affect the detection speed. Table 1 includes the time take to perform detection on single image.

\section{Conclusion}
Our evaluation and analysis gives the significant drawbacks of using greedy non-maximum suppression approach and demonstrates how it restricts the performance of conventional SSD. We also provide the effect of object size on performance of conventional SSD. This paper highlights the use of affinity propagation clustering in SSD to overcome the drawbacks of NMS. Our evaluation and analysis strongly suggest the effectiveness of proposed approach and shows its potential to provide better object detection. We also cover the application of this improvement to automate the asset-monitoring in construction sites.
\section*{Acknowledgment}

We acknowledge the support of the Natural Sciences and Engineering Research Council of Canada (NSERC), [funding reference number 396151363].
%

\bibliographystyle{IEEEtran}
\bibliography{IEEEabrv,bibliography.bib}

\end{document}